\pdfoutput=1
\documentclass[lettersize,journal]{IEEEtran}
\usepackage{amsmath,amsfonts}
\usepackage{algorithm}
\usepackage{algpseudocode}
\usepackage{array}
\usepackage[caption=false,font=normalsize,labelfont=sf,textfont=sf]{subfig}
\usepackage{textcomp}
\usepackage{stfloats}
\usepackage{url}
\usepackage{verbatim}
\usepackage{graphicx}
\usepackage{cite}

\usepackage[colorlinks,
            linkcolor=blue,
            anchorcolor=blue,
            citecolor=blue]{hyperref}
\newtheorem{theorem}{Theorem}

\newtheorem{corollary}{Corollary}

\newcommand{\R}{\mathbb{R}}

\begin{document}

\title{Stabilizing black-box algorithms through task-oriented randomization}

\author{Yali Wang and Zhaojun Wang}


\maketitle

\begin{abstract}
As black-box models become foundational to modern research, ensuring their stability is paramount for the realization of trustworthy artificial intelligence. The inherent diversity of inputs—ranging from structured Gaussian distributions to complex data with unknown structures—poses a significant challenge: how to stabilize black-box outputs while effectively leveraging available prior information. This paper introduces a task-oriented randomization methodology that adaptively tailors its strategy to the underlying generative mechanisms of the input data, specifically addressing unstructured complexities. A comprehensive suite of stability guarantees is proposed. Beyond establishing rigorous theoretical foundations for stability, the research provides a detailed analysis of the intrinsic trade-off between stability and exploration. Motivated by the architecture of Large Language Models, the framework is further extended to top-$k$ ranking problems. The validity and effectiveness of the proposal are demonstrated through extensive numerical simulations and applications to the real-world dataset.
\end{abstract}


\section{Introduction}
    \IEEEPARstart{T}{he} analysis of black-box models has become a cornerstone of the modern artificial intelligence (AI) era. As neural networks grow in complexity, their internal logic becomes increasingly opaque, rendering traditional parametric and non-parametric analytical methods insufficient for full interpretation. This lack of transparency has necessitated the development of new research pathways designed specifically to probe and demystify complex algorithmic behaviors.

    As the field of statistics evolves to meet these modern demands, the focus has shifted from pure predictive power toward veridicality. According to the framework proposed by \cite{yu2020veridical}, reliable data science must be grounded in the PCS (Predictability, Computability, and Stability) principles. Among these, stability—the requirement that results remain robust against reasonable perturbations in data and modeling choices—is considered paramount for ensuring trustworthy AI.
    
    Recent advancements have sought to operationalize this stability within black-box environments. Specifically, \cite{soloff2024Bagging}, \cite{soloff2024building}, \cite{liang2025assumption}, \cite{soloff2024stability} have introduced methods to stabilize algorithmic outputs by leveraging bagging techniques. Building on this foundation, the authors extended these methodologies to address complex top-k ranking problems. Because this framework is inherently assumption-free, it offers a versatile and straightforward solution that can be applied across a wide range of scientific and industrial domains.

    Notwithstanding its widespread application, the framework exhibits certain shortcomings when dealing with complex datasets. First, standard resampling schemes often fail to account for the underlying data structures; as a result, the drawn samples cannot reflect the true data distribution, thereby introducing bias. Second, effective sampling requires prior knowledge of the distribution, which is often unattainable in practical environments. Another naive route would be adopting a sample-splitting approach where one subset is used for distribution estimation and the other for sampling. However the inherent complexity of the data leads to poor estimation performance, while the partitioning itself results in substantial information loss, undermining subsequent sampling and inference. 

    In light of the aforementioned considerations, we are motivated to propose an alternative methodology to address the stabilization of black-box algorithms. Unlike the existing general framework, our approach is inherently input-driven, enabling it to adapt effectively to diverse data types, particularly those characterized by unknown complexities. We characterize this property as 'task-oriented'; specifically, the implementation of the proposed method is contingent upon the characteristics of the input data.

    \subsection{Challenges and Contributions}
    The primary challenge resides in managing data complexity, where the underlying distributions are often both unknown and highly intricate. The complex nature of these distributions, compounded by the inherent opacity of black-box algorithms, presents a significant obstacle to achieving reliable stabilization.
    
    Achieving assumption-lean or distribution-free properties is also essential for reliable stabilization. While traditional resampling—such as the bootstrap or sub-sampling—offers a potential path, both suffer from the aforementioned limitations, necessitating some alternatives.
    
    Beyond the challenges posed by complex data and assumption-free pursuit, there remains the question of efficiency when the input follows a known, canonical distribution, such as Gaussian observations. In such instances, a critical objective is to determine how to effectively leverage this prior information to enhance performance or reduce computation burden. 

    Synthesizing these challenges, the fundamental objective remains the attainment of the 'task-oriented' stabilization target. In response to these challenges, we introduce a novel methodology designed to adapt dynamically to diverse data structures, particularly those involving intricate and unknown complexities. Through specialized configurations, our method can recover both bootstrap based and sub-sampling based stabilization algorithms as special cases. 

    We extend our proposed method to top-k ranking by leveraging the inherent scoring mechanism of Large Language Models. Within this paradigm, the Transformer serves as a black-box scoring function, where the output values are sorted to produce the final k-ranked sequence.

    We establish formal stability guarantees for the proposed framework and provide a discussion of the inherent trade-off between system stability and exploratory behavior. We substantiate our theoretical findings through extensive evaluations on both synthetic benchmarks and real-world datasets, demonstrating the practical efficacy and scalability of the method.
    
    \subsection{Problem Formulation}
    Let $\mathcal{A}$ denote a black-box algorithm and $\mathcal{D} = \{Z_i\}_{i=1}^n$ represent an input dataset of size $n$, where each $Z_i$ is an individual observation. The operational framework of a black-box algorithm can be formally represented as the mapping of a dataset $\mathcal{D}$ through an algorithmic function $\mathcal{A}$ to yield a specific output:
    \begin{center}
	Dataset $\mathcal{D}$ \quad $\longrightarrow$ \quad Algorithm $\mathcal{A}$ \quad $\longrightarrow$\quad Output $O$.
    \end{center} 
    Despite the efficacy of such models, the output of $\mathcal{A}$ could be characterized by significant instability, demonstrating high sensitivity to minor perturbations within the input data. This lack of consistency poses a challenge to the reliability and reproducibility of the system. Consequently, the objective of this research is to implement methodologies to stabilize the output of the black-box algorithm $\mathcal{A}$, thereby ensuring more robust and predictable performance.
        
    \subsection{Intuition}\label{Intro_Intuition}
    The fundamental concept of our approach is encapsulated by the term 'noisify'. The procedural framework is structured as follows: 
    \begin{center}
	Dataset $\mathcal{D}$ \, $\rightarrow$ \, Noisified $\widetilde{\mathcal{D}}_1,\cdots,\widetilde{\mathcal{D}}_B$ \, $\rightarrow$ \, Algorithm $\mathcal{A}$ \, $\rightarrow$\, Outputs $\widetilde{O}_1,\cdots,\widetilde{O}_B$ \, $\rightarrow$ \, Output $O$
    \end{center}
    where $B$ is the number of generated nosified datasets. After input the $B$ nosified datasets, we get $B$ outputs $\widetilde{O}_1,\cdots,\widetilde{O}_B$. The final output is obtained by ensembling the results $\widetilde{O}_1,\cdots,\widetilde{O}_B$ through averaging and majority voting procedures.

    The primary challenge lies in how to noisify the dataset $\mathcal{D}$. Our objective is to develop a proposal capable of handling complex data through an assumption-lean or distribution-free approach, while remaining flexible enough to incorporate prior knowledge when available. One straightforward approach involves noise injection, with Gaussian and Laplace noises being the most common choices. However, this raises a critical challenge: determining the appropriate noise type for complex data with an unknown underlying structure. 

    Another intuitive approach would be to perturb the outputs of algorithm $\mathcal{A}$. However, because $\mathcal{A}$ is a black box, we lack fundamental knowledge regarding its internal logic and output distribution. Without understanding these characteristics—specifically the sensitivity of the output—simply adding noise or sub-sampling is insufficient to maintain the structure of the output.

    Integrating these considerations, we develop a specialized 'noisification' framework that handles complex data through an assumption-lean approach. This methodology maintains the flexibility to incorporate prior knowledge whenever it is available.

    \subsection{Road-map}
    The remainder of this paper is organized as follows. Section~\ref{Task-Oriented_Randomization} introduces our proposed methods, which are subsequently extended to the Top-K ranking problem in Section~\ref{Extension_Top-K}. In Section~\ref{TG}, we provide the theoretical guarantees for our approach, specifically focusing on stability and exploration. We then evaluate the performance of our proposal through simulations and real-data analysis in Section~\ref{Simulation} and Section~\ref{RealWorld_Data_Analysis}. Finally, Section~\ref{Works_Review} provides a comprehensive review of related work, and Section~\ref{Discussion} concludes the paper with a discussion on future research directions.

    \section{Task-Oriented Randomization}\label{Task-Oriented_Randomization}
    In Section~\ref{Intro_Intuition}, we introduce our core idea to stabilize the output of black-algorithms. Given the input data set $\mathcal{D}$, we generate $B$ nosified datasets $\widetilde{\mathcal{D}}_1, \cdots, \widetilde{\mathcal{D}}_B$. Input those nosified datasets into the black-box algorithm $\mathcal{A}$, we get the corresponding outputs $\widetilde{O}_1,\cdots,\widetilde{O}_B$. By averaging or major voting, we get the final output $O$. Our main focus would be, how to noisify the dataset $\mathcal{D}$. Rather than to be a general method, our method will be 'task-oriented', i.e., the concrete techniques are contingent upon the inherent properties of the input data.
    
    The proposed approach addresses two distinct conditions based on the transparency of the input data: (i) cases involving a prespecified generation mechanism, and (ii) cases where the underlying mechanism is unobserved.
    
    \subsection{Case 1: Generation Mechanism of the Input Data, Known}
    Suppose we know the generation process of the input data, then we generate random noises $\varepsilon_1, \cdots, \varepsilon_B$. The corresponding noisified datasets $\widetilde{\mathcal{D}}_1, \cdots, \widetilde{\mathcal{D}}_B$ would be
    \begin{equation}
        \widetilde{\mathcal{D}}_i = \{ Z_j + \varepsilon_{i,j} \}_{j=1}^n, \quad\text{for }1\leq i\leq B.
    \end{equation}
    The generated random noises $\varepsilon_1, \cdots, \varepsilon_B$ should be chosen according to the generation process of the input data. For example, if we know the input data should be normal distributed, then the add noises can be chosen as the Gaussian ones. Other widely known choices are the Laplace noises, the Exponential noises. Which one to choose will depends on the generation mechanism of the input data.

    We can also recover the bootstrap and sub-sampling by choosing specific noises. For $1\leq i\leq B$, 
    \begin{itemize}
        \item Bootstrap : $\varepsilon_{i,j}$ are drawn randomly with replacement from $\{ Z_k-Z_j: 1\leq k\leq n \}$,
            \begin{equation}
            \widetilde{\mathcal{D}}_i = \{ Z_j + \varepsilon_{i,j} \}_{j=1}^n.
            \end{equation}
        \item Sub-sampling : $\varepsilon_{i,j}$ are drawn randomly without replacement from $\{ Z_k-Z_j: 1\leq k\leq n \}$,
            \begin{equation}
            \widetilde{\mathcal{D}}_i = \{ Z_j + \varepsilon_{i,j} \}_{j=1}^m,
            \end{equation}
        where $m<n$ is a pre-specified number of draws.
    \end{itemize}
    In comparing bootstrap and sub-sampling methods, \cite{politis1999subsampling} characterize the bootstrap as a 'right size, wrong distribution' approach, while sub-sampling is described as a 'wrong size, right distribution' method. This distinction highlights that while the bootstrap uses the full sample size $n$, it relies on the empirical distribution which may be invalid for certain statistics; conversely, subsampling uses a reduced block size $m < n$ to maintain the validity of the underlying distribution.

    \subsection{Case 2: Generation Mechanism, Unknown}
    Resampling methods, such as sub-sampling and bootstrapping, can be inefficient when the underlying data generation process is unknown—particularly when dealing with complex or poorly understood datasets. 
    
    To address the aforementioned limitations, we adopt an alternative approach: the diffusion process. This framework operates on a simple premise: by modeling both the forward degradation of data into noise and the subsequent reverse process, we can reconstruct complex, high-fidelity data from a stochastic, noisy distribution. 

    The intuition behind the diffusion process lies in its two-phase architecture. During the forward process (or 'diffusion'), a dataset is systematically transformed into Gaussian white noise by adding controlled increments of variance. Because each step of noise addition is known, we can train a model to approximate the backward process (or 'reverse diffusion'). By mastering this reversal, the model learns to iteratively remove noise, allowing it to synthesize high-quality target data from a purely stochastic starting point.

    The efficiency of the diffusion model is not merely theoretical; it is validated by a growing body of research and widespread industrial adoption. And this process powers the most advanced AI technologies in the world. For example, frontier models such as Sora applies diffusion principles to create temporally consistent, high-definition video.

    Now we mathematically formalize our methods. 
    \begin{itemize}
        \item \textsl{The Forward Diffusion Process.} At each time step $t\in\{ 1,\cdots, T\}$, the transition kernel is defined by a Gaussian distribution: 
        \begin{align}
            q(z_t|z_{t-1})=N(z_t;\sqrt{\alpha_t}z_{t-1},\beta_tI)
        \end{align}
        where $\beta_t$ represents the predefined noise schedule, $\alpha_t=1-\beta_t$, $I$ is the identity matrix. Then the closed-form expression for the marginal distribution $q(z_t|z_{0})$ at any arbitrary time $t$. The marginal distribution is given by:
        \begin{align}
            q(z_t|z_{0})=N(z_t;\sqrt{\bar{\alpha}_t}z_{0},(1-\bar{\alpha}_t)I)
        \end{align}
        where $\bar{\alpha}_t=\prod_{i=1}^t\alpha_i$.
        \item \textsl{The Reverse Generation Process.} Given the pure isotropic Gaussian noise $z_T\sim N(0,I)$, aim to the reverse distribution. Train a network $p_{\theta}$ to approximate
        \begin{align}
            p_{\theta}(z_{t-1}|z_t)=N(z_{t-1};\mu_{\theta}(z_t,t),\Sigma_{\theta}(z_t,t)).
        \end{align}
    \end{itemize}
    Perform like this, we can get several diffused datasets $\widetilde{D}_1,\cdots,\widetilde{D}_B$.

    \subsection{Algorithm}
    The proposed procedure is detailed in Algorithm~\ref{Alg:nosification}. To synthesize the final result from the individual outputs $\{\widetilde{O}_b\}_{b=1}^B$, we employ an aggregation function. While this function can be tailored to the specific application—for instance, using majority voting for discrete labels or the mean for continuous values—this paper focuses on the averaging approach. A more detailed exploration of alternative aggregation strategies is provided in the Discussion section.
    
\begin{algorithm}[t]
    \caption{\textbf{Noisification Algorithm $\mathcal{A}^{Nos}$}}
    \label{Alg:nosification}
    
    \begin{algorithmic}
    \renewcommand{\algorithmicrequire}{\textbf{input}}
    \renewcommand{\algorithmicensure}{\textbf{output}}
    
    \Require Base algorithm $\mathcal{A}$; data set $\mathcal{D}$ with $n$ training points; $B$

    \For {$b = 1, \dots, B$}
        \State Generate $B$ noisifed data sets $\widetilde{\mathcal{D}}_b$ according to the scenarios
        
       \State Input the $\widetilde{\mathcal{D}}_b$ into the algorithm $\mathcal{A}$ and get the corresponding output $\widetilde{O}_b$
    \EndFor
    
    \Ensure $O$, which combines the outputs $\widetilde{O}_b$ for $b=1,\cdots,B$ by averaging function
        \begin{equation}
         O=\frac{1}{B}\sum_{b=1}^B\widetilde{O}_b
    \end{equation}
    \end{algorithmic}
\end{algorithm}

\section{Extension to Top-K Ranking Problem}\label{Extension_Top-K}

    Ranking problems are ubiquitous across various machine learning domains, notably in recommendation systems and sequential decision-making frameworks. Recently, Large Language Models (LLMs) leveraging the Transformer architecture have emerged as proficient "prediction engines" for these tasks. These models operate by estimating the conditional probability distribution over a discrete vocabulary $\mathcal{V}$ to determine the most likely succeeding token. This mechanism can be formally decomposed into a two-stage process:
    \begin{itemize}
        \item Scoring (Logit Generation): The model maps high-dimensional hidden representations to a set of raw scalar values, or logits, which quantify the relative affinity for each candidate token in the vocabulary based on the preceding context.
        \item Decoding (Selection Strategy): The model applies a stochastic or deterministic sampling heuristic—such as Top-$k$ sampling—to select a discrete token from the transformed probability distribution.
    \end{itemize}

    The proposed framework can be generalized to this setting by treating the Transformer architecture as a black-box algorithm, where $\widetilde{O}_b$ represents the vector of the top-$k$ scores (logits) generated for the $b$-th input instance. Following $B$ independent stochastic passes, a majority-voting consensus mechanism is applied to aggregate the results. This procedure identifies the most statistically robust tokens from the output distribution, effectively leveraging the ensemble of trials to select the optimal candidates.

\section{Theoretical Guarantees}\label{TG}
    To provide a rigorous foundation for task-oriented randomization, we establish a formal theoretical framework. We first derive stability guarantees in Section \ref{TG_SG} and then presents a detailed analysis of the exploration ability in Section~\ref{TG_EA}.

    \subsection{Stability Guarantees}\label{TG_SG}
        \subsubsection{$(s,\delta)$-Stability}
    We first follow the stability formulation proposed by \cite{bousquet2002stability, elisseeff2005stability} to give our stability definition. An algorithm $\mathcal{A}^{Nos}$ is $(s,\delta)$-stable if, for the training data set $\mathcal{D}=(Z_i)_{i=1}^n$ and test point $z$,
    \begin{align}
    \frac{1}{n}\sum_{i=1}^n\mathbb{P}\left\{ \left| \mathcal{A}^{Nos}(\mathcal{D}) (z) - \mathcal{A}^{Nos}(\mathcal{D}^{\setminus i})(z) \right|>s \right\} \leq \delta.
    \end{align}  
    
    Theorem~\ref{TG_SG_1} give the $(s,\delta)$-stability guarantee for the algorithm $\mathcal{A}^{Nos}$. 
    
    \begin{theorem}\label{TG_SG_1}
        For any algorithm $\mathcal{A}$, the noisified algorithm $\mathcal{A}^{Nos}$ is $(s,\delta)$-stable if 
        \begin{align}
            \delta s\geq\sqrt{\mathcal{E}_n/n}
        \end{align}
        where $\mathcal{E}_n=\sum_{i\leq n}\left( \mathbb{E}_b\left\{ \mathcal{A}(\widetilde{\mathcal{D}}_b)(z) - \mathcal{A}(\widetilde{\mathcal{D}}_b^{\setminus i})(z) \right\} \right)^2$, with $\widetilde{\mathcal{D}}_b^{\setminus i}$ are generated through removing the $i$-th observation in $\widetilde{\mathcal{D}}_b$. Here, we take the expectation $\mathbb{E}_b$ over the randomness of the mechanism to provide a deterministic value. if the algorithm $\mathcal{A}$ depends on some random parameter $\xi$, then the $\mathcal{E}_n$ will equal to $\sum_{i\leq n}\left( \mathbb{E}_b\left\{ \mathcal{A}(\widetilde{\mathcal{D}}_b;\xi)(z) - \mathcal{A}(\widetilde{\mathcal{D}}_b^{\setminus i};\xi)(z) \right\} \right)^2$.
    \end{theorem}
    
    The term $\mathcal{E}_n$ quantifies the inherent randomness introduced by the noisification mechanism. Intuitively, a higher noise intensity increases the randomness of the mechanism. This makes it strictly more difficult to achieve $(s,\delta)$-stability, as the algorithm's output becomes more sensitive to individual stochastic perturbations. Conversely, stability is positively correlated with the number of generated noisified datasets $B$. As $B$ increases, $\mathcal{E}_n$ tends to decrease as the aggregated output tends to stabilize the averaging effect across multiple noise realizations.

    The following corollaries describe some specific cases.
    \begin{corollary}\label{coro1}
        Assuming the output of algorithm $\mathcal{A}$ is bounded within the interval $[0,1]$, the noisified algorithm $\mathcal{A}^{Nos}$ satisfies $(s, \delta)$-stability provided that the condition $\delta s \geq 1$ holds. 
    \end{corollary}

    \begin{corollary}\label{coro2}
        Assuming the output of algorithm $\mathcal{A}$ is bounded within the interval $[a,b]$, the noisified algorithm $\mathcal{A}^{Nos}$ satisfies $(s, \delta)$-stability provided that the condition $\delta s\geq |b-a|$ holds.  
    \end{corollary}

        \subsubsection{Stability Measured by Prediction Error}  
    Besides the established stability framework proposed by \cite{bousquet2002stability, elisseeff2005stability}, we also consider another measure of the stability of the algorithm $\mathcal{A}^{Nos}$. We evaluate the prediction discrepancy between the original black-box algorithm, $\mathcal{A}(\mathcal{D})$, and our proposed framework, $\mathcal{A}^{Nos}(\mathcal{D})$. 
    \begin{theorem}\label{TG_PE} For the algorithm $\mathcal{A}$ and its corresponding noisified algorithm $\mathcal{A}^{Nos}$, the training set $\mathcal{D}$ and the test point $z$, we assume the output of $\mathcal{A}$ is bounded within $[a,b]$. Denote $m^{Nos}(z)=\mathbb{E}[\mathcal{A}^{Nos}(\mathcal{D})(z)]$ where the expectation taken over the randomization of $B$ outputs. Denote $\delta(z) = |\mathcal{A}(\mathcal{D})(z) - m^{Nos}(z) |$. Then for $\varepsilon\geq\delta(z)$,
        \begin{align}
        \mathbb{P}\left( \left| \mathcal{A}(\mathcal{D})(z) - \mathcal{A}^{Nos}(\mathcal{D})(z) \right| > \varepsilon \right) \leq 2\exp\left( - \frac{2(\varepsilon - \delta(z))^2}{B(b-a)^2} \right).
        \end{align}
    \end{theorem}

     \subsubsection{Discussion of Lipschitz ratio}  
    The Lipschitz ratio also serves as a rigorous metric for evaluating the local sensitivity of the proposed algorithm, $\mathcal{A}^{\text{Nos}}$. Formally, for any two test points $z$ and $z'$, the ratio is defined by the limit:
    \begin{align}
        lim_{z'\to z}\frac{\left| \mathcal{A}^{Nos}(\mathcal{D})(z) - \mathcal{A}^{Nos}(\mathcal{D})(z') \right|}{|z-z'|}.
    \end{align}
    Given that $\mathcal{A}$ operates as a black-box function, direct analytical derivation is often infeasible. Consequently, we recommend an empirical estimation approach to approximate the local Lipschitz constant in practice.
    
    In the subsequent analysis, we investigate the exploratory capacity induced by the noisification process. While the previous section established that noise complicates the achievement of $(s, \delta)$-stability, it is important to recognize that noise is not merely a source of variance, but a mechanism for broader search space coverage.

    \subsection{Exploration Analysis}\label{TG_EA}
    In conventional statistical analysis, noises will be viewed as the tradeoff between the bias and the variance. More noises, then the estimation will become more inaccurate and the output can be biased. Fewer noises, the estimation will become more accurate, then the output will performs well. 

    However, the above discussion, depends on the stability of the learning algorithm. As to the black-box algorithm, sensitivity can be a problem. In this case, we no longer view noises as the trade-off between bias and variance. Rather, we'll view the noises as the tool to explore new frontiers. After noisification, the input data will have more possibilities, i.e., the scope of input has pushed into broader territories. So in our analysis, we view the noisy extent, as the "exploration" measure. This also brings the trade-off problem between the stability and the exploration.

    For the first case that the generation mechanism of the input data is known, the exploration measure can be taken as the variance of the added noises. For the second case that the generation mechanism of the input data is unknown, we take the Theorem 2 in \cite{chen2022sampling} as an example. The main result can be sketched as: Denote $p_T$ to be the output of the reverse generated process after $T$ steps and $q_0$ to be the true distribution of the dataset. Under some conditions, the total variation distance between $p_T$ and $q_0$ can be split into three terms: convergence of forward process, discretizations error, score estimation error. Then in the second case, we will take the total variance as the exploration measure. Both the variance or the total variance are large, then the exploration power of the mechanism is stronger.

\section{Simulation}\label{Simulation}
    We first provide a comparative visualization to demonstrate the relative stability improvements of the proposed $\mathcal{A}^{Nos}$ algorithm over the baseline algorithm $\mathcal{A}$. Then we evaluates the impact of increased noise on $(s,\delta)$-stability and prediction error. 
    
    The simulation setting follows the work \cite{soloff2024Bagging}. The observations are generated through the following data generating process:
    \begin{align}
        & X_i\stackrel{i.i.d.}{\sim} N(0,I_d),\\
        & Y_i|X_i\stackrel{ind.}{\sim}Bernoulli\left( \frac{1}{1+exp(-X_i^T\theta^*)} \right).
    \end{align}
    $d$ is taken to be 200, and $\theta^*$ is taken to be $(0.1,\cdots,0.1)\in\mathbb{R}^d$. The sample size for the training set is set to be $n_{train}=2000$ and the testing set is set to be $n_{test}=500$. For convenience, we denote the training set to be $\mathcal{D}_{train}=\{ (X_i,Y_i)\}_{i\leq n_{train}}$ and the testing set to be $\mathcal{D}_{test}$. The algorithm of $\mathcal{A}$ will be chosen as the neural network with a single hidden layer or the $L2$-regularized logistic regression that for the testing point $x$, $f_{\hat{\theta}}(x):=\left( 1+e^{-x^T\hat{\theta}} \right)^{-1}$, with $\hat{\theta}$ calculated from $\mathcal{D}$, through
	\begin{equation}
        \begin{split}
		\hat{\theta}=argmin_{\theta\in\R^d} & \left\{ C\sum_{i=1}^n\left( -Y_i\log(f_{\theta}(X_i)) \right.\right.\\
        & \left.\left.- (1-Y_i)\log(1-f_{\theta}(X_i)) \right) + \frac{1}{2}\| \theta \|_2^2 \right\}.
        \end{split}
	\end{equation}
    Comprehensive details regarding parameter selection are provided in the Appendix. 

    We take $B=1000$. For each $b \in \{1, \dots, B\}$, the noisified dataset is generated by:
    \begin{align}
    \widetilde{\mathcal{D}}_b^{\sigma} = \{ (X_i + \varepsilon_i^b, Y_i) \}_{i=1}^{n_{train}},
    \end{align}
    where the noise terms $\varepsilon_i^b$ are independent and identically distributed samples drawn from a multivariate Gaussian distribution, $\varepsilon_i^b \sim N(0, \sigma^2 I_d)$. $\sigma\in(0,4]$. Detailed implementation of these methods is provided in the Appendix.
    
    \subsection{Comparison between $\mathcal{A}$ and $\mathcal{A}^{Nos}$}
    This section evaluates the stability of the original algorithm $\mathcal{A}$ in comparison to the proposed $\mathcal{A}^{Nos}$. Specifically, we measure the absolute difference in model outputs when a single training datum $k$ is removed: $|\mathcal{A}(\mathcal{D})(z) - \mathcal{A}(\mathcal{D}^{\setminus k})(z)|$ and $|\mathcal{A}^{Nos}(\mathcal{D})(z) - \mathcal{A}^{Nos}(\mathcal{D}^{\setminus k})(z)|$.  

\begin{figure}[htbp]
    \centering
    \includegraphics[width=3in]{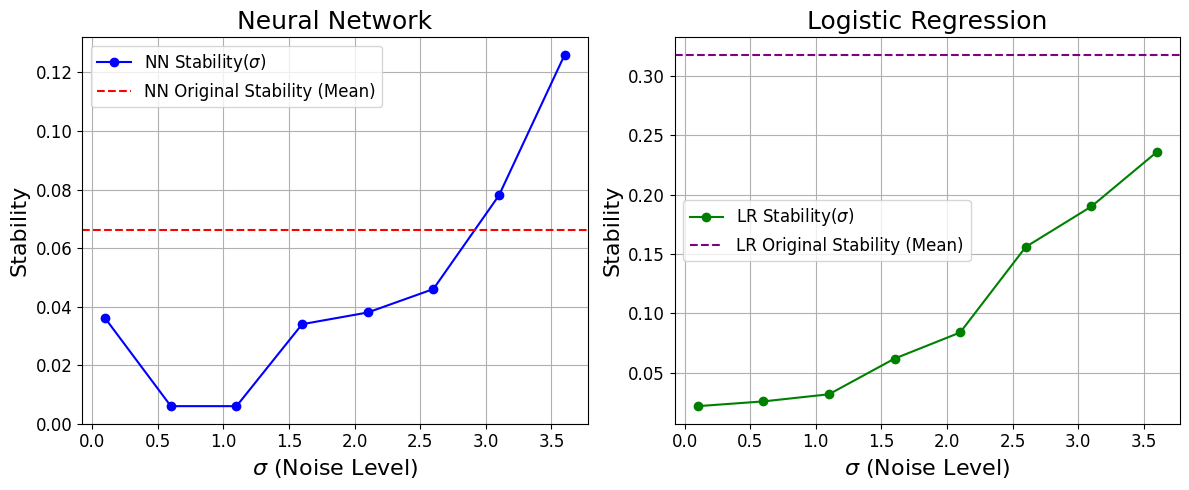}
    \caption{Stability Comparison $\mathcal{A}$ and $\mathcal{A}^{Nos}$. (Left) $\mathcal{A}$ to be the Neural Network. The blue line visualize the $|\mathcal{A}^{Nos}(\mathcal{D}) (z) - \mathcal{A}^{Nos} (\mathcal{D}^{\setminus k}) (z)|$, while the red horizontal line visualize the $|\mathcal{A}(\mathcal{D})(z) - \mathcal{A}(\mathcal{D}^{\setminus k})(z)|$. (Right) $\mathcal{A}$ to be the $L2$-regularized Logistic Regression. The green line visualize the $|\mathcal{A}^{Nos}(\mathcal{D}) (z) - \mathcal{A}^{Nos} (\mathcal{D}^{\setminus k}) (z)|$, while the purple horizontal line visualize the $|\mathcal{A}(\mathcal{D})(z) - \mathcal{A}(\mathcal{D}^{\setminus k})(z)|$.}
    \label{Comparison_Sim}
    \end{figure}

    As illustrated in Figure \ref{Comparison_Sim}, the sensitivity of the algorithm—quantified by the difference $|\mathcal{A}^{Nos}(\mathcal{D})(z) - \mathcal{A}^{Nos}(\mathcal{D}^{\setminus k})(z)|$—increases proportionally with the noise parameter $\sigma$. In the case of the neural network, we observe that for $\sigma < 2.7$, $|\mathcal{A}^{Nos}(\mathcal{D})(z) - \mathcal{A}^{Nos}(\mathcal{D}^{\setminus k})(z)|$ is less than $|\mathcal{A}(\mathcal{D})(z) - \mathcal{A}(\mathcal{D}^{\setminus k})(z)|$. This indicates that $\mathcal{A}^{Nos}$ exhibits greater algorithmic stability than the original algorithm $\mathcal{A}$ within this regime. However, when $\sigma > 2.7$, $\mathcal{A}^{Nos}$ becomes increasingly unstable. These results suggest that while a calibrated level of noise can enhance stability, excessive noise leads to significant signal degradation. Consequently, a vital direction for future research involves developing a systematic method for selecting the optimal noise level $\sigma$ to maximize the efficacy of the proposed $\mathcal{A}^{Nos}$ framework.

    In contrast to the neural network results, the logistic regression model exhibits significantly higher variability in its stability measurements. Specifically, the sensitivity of the neural network, $|\mathcal{A}^{Nos}(\mathcal{D})(z) - \mathcal{A}^{Nos}(\mathcal{D}^{\setminus k})(z)|$, remains within a narrow range of $0.01$ to $0.13$. In comparison, the logistic regression model demonstrates a wider fluctuation, with values ranging from $0.01$ to $0.25$. These observations suggest that, on average, the neural network maintains superior algorithmic stability compared to the logistic regression approach under the evaluated conditions.
    
    \subsection{Effects of Noises}
\begin{figure}[htbp]
    \centering
    \includegraphics[width=3in]{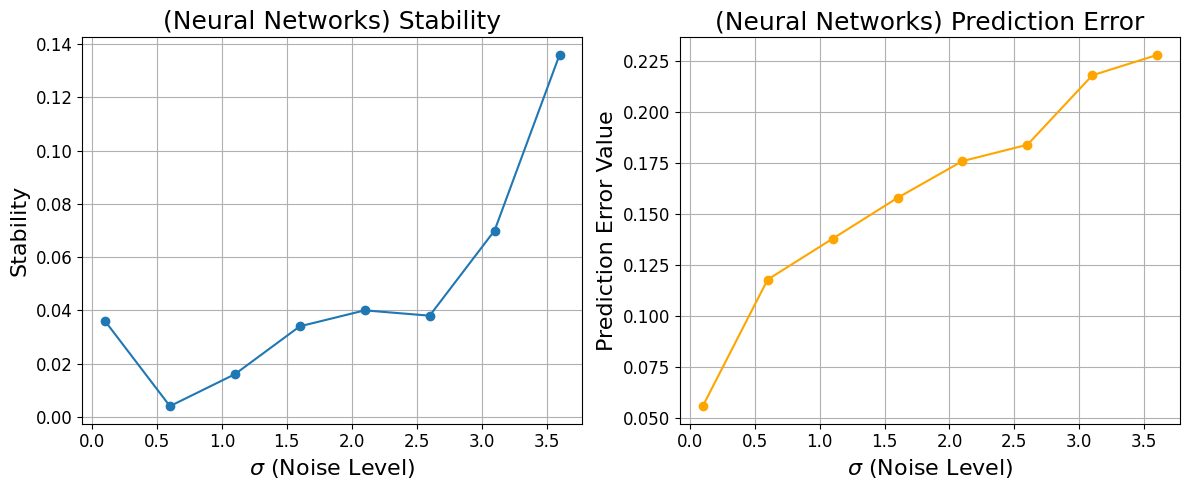}
    \caption{Stability and Prediction Error of Neural Networks. (Left) Algorithmic stability $|\mathcal{A}^{Nos}(\mathcal{D})(x) - \mathcal{A}^{Nos}(\mathcal{D}^{\setminus k})(x)|$ as a function of noise level $\sigma$. (Right) Prediction error $|\mathcal{A}(\mathcal{D})(x)-\mathcal{A}^{Nos}(\mathcal{D})(x)|$ relative to noise level $\sigma$.}
    \label{NN}
    \end{figure}

\begin{figure}[htbp]
    \centering
    \includegraphics[width=3in]{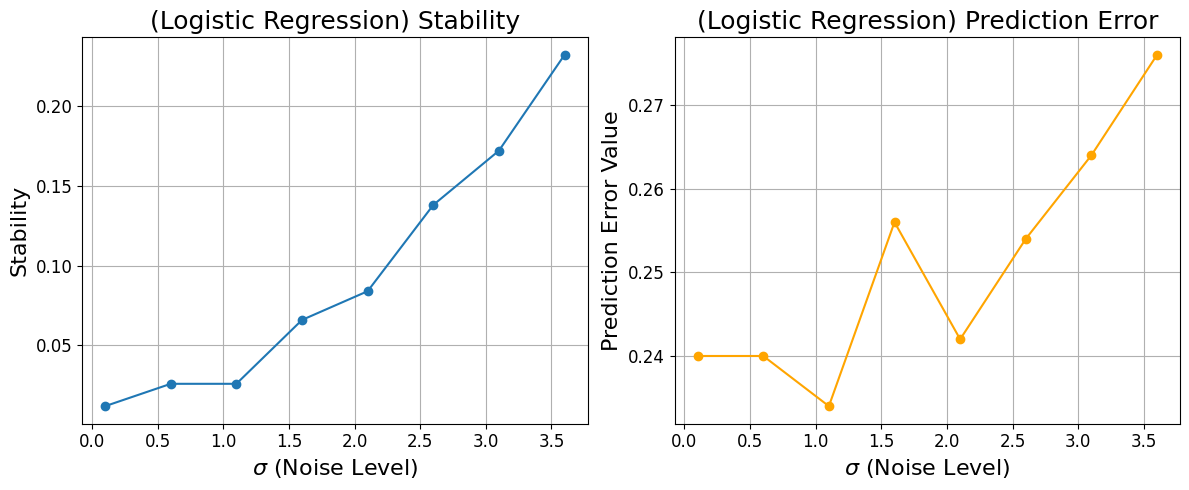}
    \caption{Stability and Prediction Error of $L_2$-regularized Logistic Regression. (Left) Algorithmic stability $|\mathcal{A}^{Nos}(\mathcal{D})(x) - \mathcal{A}^{Nos}(\mathcal{D}^{\setminus k})(x)|$ as a function of noise level $\sigma$. (Right) Prediction error $|\mathcal{A}(\mathcal{D})(x)-\mathcal{A}^{Nos}(\mathcal{D})(x)|$ relative to noise level $\sigma$.}
    \label{LR}
    \end{figure}

    This section evaluates the impact of noise by analyzing $(s, \delta)$-stability alongside prediction error. Figures~\ref{NN} and \ref{LR} illustrate these results for the neural network and $L_2$-regularized logistic regression architectures, respectively. Observations indicate that for both models, an increase in the noise parameter $\sigma$ correlates with a rise in the stability measure $|\mathcal{A}^{Nos}(\mathcal{D})(x) - \mathcal{A}^{Nos}(\mathcal{D}^{\setminus k})(x)|$, suggesting a reduction in algorithmic stability. There is also an increase in prediction error when $\sigma$ increases. This means that the difference between $\mathcal{A}(\mathcal{D})(x)$ and $\mathcal{A}^{Nos}(\mathcal{D})(x)$ enlarges.
    
\section{Real-World Data Analysis}\label{RealWorld_Data_Analysis}
    We evaluate our method using the MNIST dataset, a standard benchmark for handwritten digit recognition. It consists of 70,000 grayscale images (digits 0–9), with a predefined split of 60,000 training and 10,000 test samples. Each image is represented as a $28 \times 28$ pixel grid, forming a 784-dimensional feature vector of grayscale intensities.

    To facilitate rapid model prototyping, two reduced subsets of the MNIST dataset were derived. First, a mini-MNIST subset was established by randomly drawing 600 training and 100 test samples for each digit. From this intermediate set, we further extracted a minimini-MNIST dataset, which consists of a reduced training pool of 500 samples per class while keeping the test set consistent with the mini-MNIST configuration.

    Our experimental procedure is outlined as follows: First, for the mini-MNIST dataset, we employ a diffusion process to generate a synthetic counterpart for each original image in the training set. By augmenting the original data with these generated samples, we obtain an expanded training set of 12,000 images. A Convolutional Neural Network (CNN) is then trained on this augmented dataset to predict labels for the test set.

    To further evaluate the stability and robustness of our approach, we conduct a similar analysis on the minimini-MNIST dataset. We generate one synthetic image for each training sample via the diffusion process, resulting in a total of 10,000 training images. The same CNN architecture is utilized to perform classification and record the predictive accuracy on the test data.

    \begin{figure}[htbp]
    \centering
    \includegraphics[width=3in]{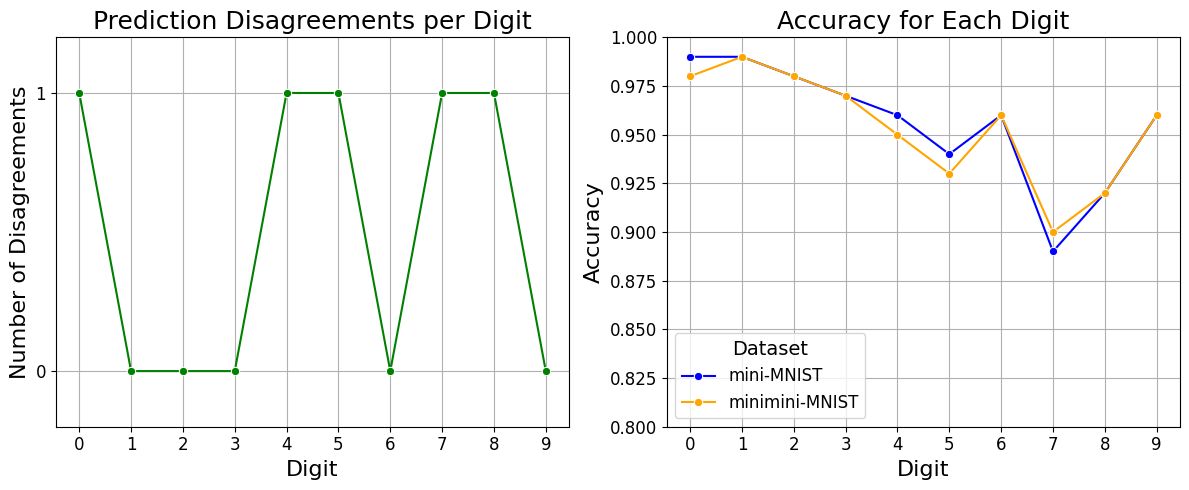}
    \caption{Evaluation of stability and classification accuracy. The left panel illustrates the number of prediction disagreements based on the mini-MNIST and on the minimini-MNIST datasets, where the x-axis represents the digit class and the y-axis shows the prediction discrepancy for each category. The right panel displays a comparative analysis of accuracy across both datasets.}
    \label{MNISTDiff_App}
    \end{figure}

    Figure~\ref{MNISTDiff_App} presents the experimental results. The left panel illustrates the stability measure by plotting the prediction discrepancy between the mini-MNIST and minimini-MNIST configurations. Notably, the disagreement count remains below 1 for all digit classes. Given that each digit consists of 100 test images, this translates to a discrepancy rate of less than 1\% (0.01), indicating a high degree of model consistency across different training scales. The right panel compares the classification accuracy of the two settings. The mini-MNIST configuration generally outperforms the minimini-MNIST one.
    
\section{Works Reviews}\label{Works_Review}
    The theoretical foundation of Algorithmic Stability as a means to guarantee generalization was largely solidified by the seminal works of \cite{bousquet2002stability} and \cite{elisseeff2005stability}. Following these foundational papers, a vast body of literature has been established to extend these concepts into various domains of machine learning. Recent studies by \cite{soloff2024Bagging, soloff2024building, liang2025assumption, soloff2024stability} have explored bagging-based algorithms to achieve "assumption-free" stability, specifically adapting these frameworks to address the ranking problems.

    Beyond classical stability, Differential Privacy (DP) has emerged as a closely related paradigm for guaranteeing generalization, with foundational principles established in the seminal works of Dwork et al. (\cite{dwork2006calibrating, dwork2014algorithmic, dwork2015preserving}). While standard DP assumes a trusted centralized curator, recent research has pivoted toward Local Differential Privacy (LDP) to mitigate risks associated with central data aggregation. LDP shifts the perturbation process to the client-side, ensuring that sensitive information remains protected even from the data collector. For a comprehensive taxonomy of LDP mechanisms and their applications in machine learning, we refer the reader to the survey by \cite{yang2024local}.

    The theoretical foundations and practical applications of the bootstrap were pioneered by Bradley Efron in his seminal works \cite{efron1986bootstrap, efron1992bootstrap}. To address scenarios where standard resampling may prove inconsistent, Politis and Romano introduced the sub-sampling framework, providing key developments in both general large-sample theory \cite{politis1994large} and applications to stationary time series \cite{politis1994stationary}. Comprehensive overviews of these resampling paradigms can be found in the fundamental monographs by Efron \cite{efron1982jackknife, efron2021computer}.
    
    The literature on diffusion processes is extensive; however, we constrain our focus to the foundational frameworks that define the field. Historically, deep generative modeling has been characterized by a persistent tension between two competing objectives: the flexibility required to characterize complex data distributions and the tractability necessary for efficient learning, sampling, and posterior evaluation. Diffusion Probabilistic Models have emerged as a robust solution to this trade-off, utilizing a strategy deeply rooted in non-equilibrium statistical physics. The theoretical architecture of modern DMs was pioneered by \cite{sohl2015deep}, who demonstrated that principles from non-equilibrium thermodynamics could be leveraged to design generative algorithms that are both stable to train and analytically tractable through an iterative reversal of the diffusion process.
    
    The widespread adoption and success of diffusion models are largely attributable to a series of pivotal architectural and methodological refinements that mitigated early computational bottlenecks and introduced sophisticated control mechanisms (\cite{ho2020denoising, song2020denoising, dhariwal2021diffusion, rombach2022high, song2020score}).

\section{Discussion}\label{Discussion}
    In this study, we introduced a novel framework for stabilizing black-box algorithms via task-oriented randomization. In contrast to traditional ensemble methods like bagging, our approach provides a more versatile mechanism for handling heterogeneous input types and high-dimensional, complex data. Beyond empirical performance, we established theoretical analysis for both stability and exploration. 

    Our current analysis focuses on a fundamental definition of stability. However, the literature distinguishes between various degrees of sensitivity, most notably the point-wise and uniform variations established by \cite{elisseeff2005stability}. Transitioning from general stability to these more stringent theoretical guarantees would provide a more granular understanding of how specific data points influence model behavior.
    
    In this study, we employed an averaging function to aggregate the outputs $\widetilde{O}_1,\cdots,\widetilde{O}_B$. However, this aggregation strategy can be generalized to include majority voting, median selection, or other robust estimators, depending on the specific task requirements and the statistical properties of the output space. The extension to these alternative functions would necessitate the development of new theoretical guarantees, which remains a promising direction for future work.
 
    We have provided an initial extension of our framework to the Top-$K$ Sampling problem. However, a rigorous exploration of the theoretical guarantees, algorithmic implementation details, and empirical performance in this context is still required. Investigating how stability behaves in structured output spaces—where the "best" result is a ranked list rather than a single value—represents a significant opportunity for future inquiry.

\bibliographystyle{abbrvnat}
\bibliography{reference}

\end{document}